\documentclass[10pt,twocolumn,letterpaper]{article}
\pdfoutput=1
\usepackage[dvipdfmx]{graphicx}
\usepackage{cvm}
\usepackage{times}
\usepackage{epsfig}
\usepackage{amsmath}
\usepackage{amssymb}

\usepackage[pagebackref=true,breaklinks=true,letterpaper=true,colorlinks,bookmarks=false]{hyperref}
\usepackage{bmpsize}

\cvmfinalcopy % *** Uncomment this line for the final submission

\def\cvmPaperID{210} % *** Enter the cvm Paper ID here

\ifcvmfinal\fi

\usepackage{amsmath,amsthm,amssymb}
\usepackage{graphicx}
\usepackage{textcomp}
\usepackage{wrapfig}
\usepackage{subfig}
\usepackage{color}
\usepackage{xspace}
\usepackage{overpic}
\usepackage{subfig}
\usepackage{enumitem}
\usepackage{booktabs}
\usepackage{multirow}
\usepackage{color}
\definecolor{blue}{rgb}{0,0,1}
\definecolor{red}{rgb}{1,0,0}
\definecolor{green}{rgb}{0,.5,0}
\definecolor{orange}{rgb}{0.75, 0.4, 0}

\newcommand{\rev}[1]{{\color{black}\textbf{}#1}\normalfont}

\newcommand{\shortcite}{\cite}

\begin{document}

\title{VGF-Net: Visual-Geometric Fusion Learning \\ for Simultaneous Drone Navigation and Height Mapping}

\author{Yilin Liu\\
	Shenzhen University 
	\and 
	Ke Xie\\ 
Shenzhen University 
	\and 
	Hui Huang\thanks{Corresponding author: Hui Huang (hhzhiyan@gmail.com)}\\ 
	Shenzhen University }

\maketitle

\begin{abstract}
The drone navigation requires the comprehensive understanding of both visual and geometric information in the 3D world. In this paper, we present a \emph{Visual-Geometric Fusion Network} (VGF-Net), a deep network for the fusion analysis of visual/geometric data and the construction of 2.5D height maps for simultaneous drone navigation in novel environments. Given an initial rough height map and a sequence of RGB images, our VGF-Net extracts the visual information of the scene, along with a sparse set of 3D keypoints that capture the geometric relationship between objects in the scene. Driven by the data, VGF-Net adaptively fuses visual and geometric information, forming a unified \emph{Visual-Geometric Representation}. This representation is fed to a new \emph{Directional Attention Model} (DAM), which helps enhance the visual-geometric object relationship and propagates the informative data to dynamically refine the height map and the corresponding keypoints. An entire end-to-end information fusion and mapping system is formed, demonstrating remarkable robustness and high accuracy on the autonomous drone navigation across complex indoor and large-scale outdoor scenes.
\end{abstract}

%\keywords{drone navigation, height mapping, visual representation, geometric representation, visual-geometric fusion.}

% Use the [Name Year] citation style as required by Siggraph
%\citestyle{acmauthoryear}

%%% This is the ``teaser'' command, which puts an figure, centered, below
%%% the title and author information, and above the body of the content.

%\input{teaser}
%\begin{teaserfigure}
%	\centering
%	\includegraphics[width=1.0\linewidth]{Images/teaser_kitten}
%	\caption{Reconstructing a transparent kitten. For each view angle, the right image is the rendering of the ground truth, and the left one is our recovery model.}
%	%	\mlc{the last two figures have different view angles as the first two.  Can we rotate them a bit more to the left?}}
%	%	\vspace{-0.15in}
%	\label{fig:teaser}
%\end{teaserfigure}

% 50 word summary
%\conceptlist
%\printcopyright

\section{Introduction}

\label{sec:intro}
\begin{figure}[t!]
	\centering
	\includegraphics[width=\linewidth]{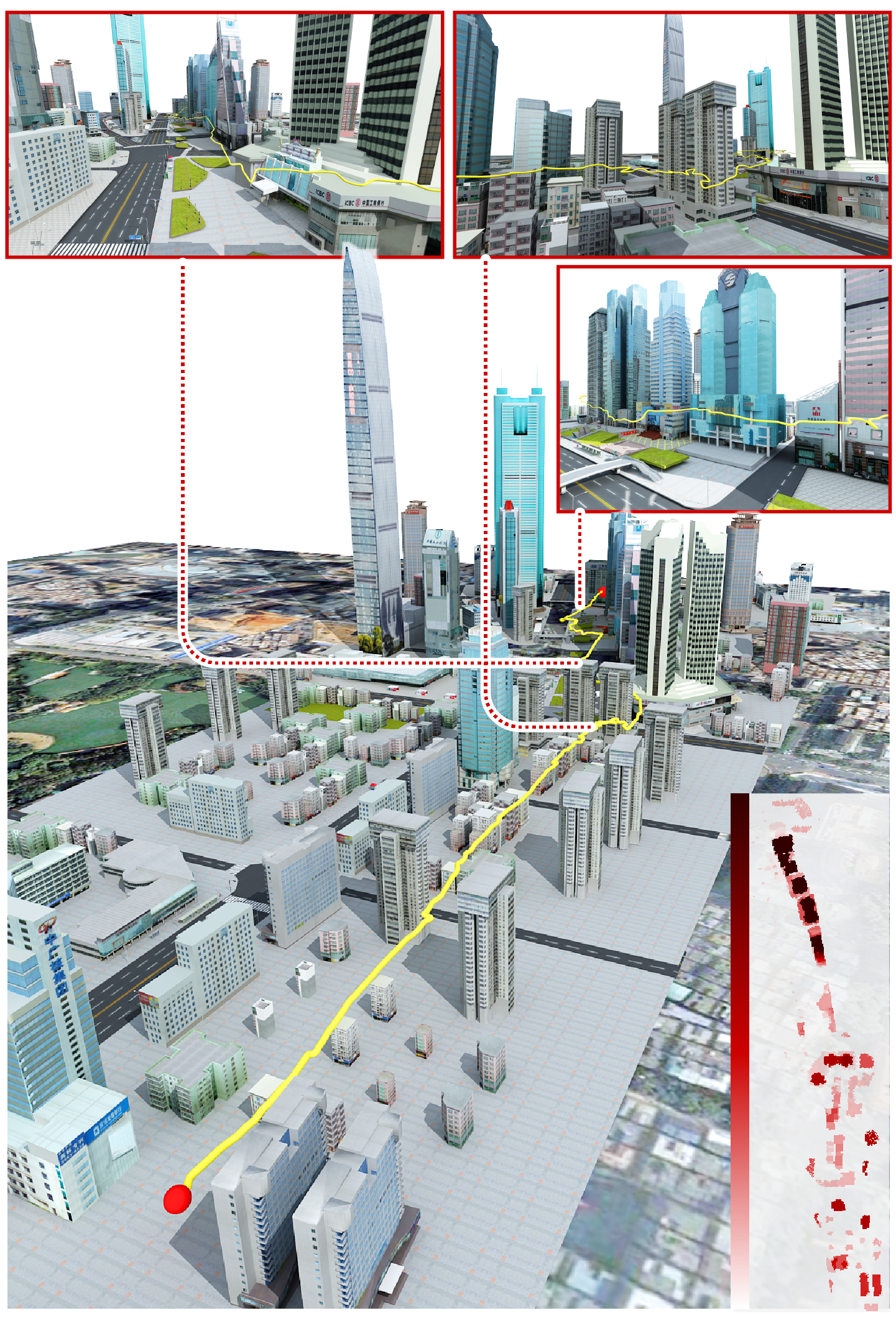}
	\caption{We show a drone navigation trajectory (yellow curve) in 3D scene, which connects the starting and target points (red dots). During the navigation, our VGF-Net dynamically updates the 2.5D height map (see the bottom-left corner) in new places (see pictures in red rectangles), which is used to timely update the navigation trajectory.}
	\label{fig:teaser}
\end{figure}

% the importance of mapping?
In recent years, we have witnessed the development of autonomous robotic systems that have been broadly used in many scenarios (e.g., autonomous driving, manufacturing and surveillance). Drone belongs to the robotic system, and is well-known for its flying capacity. Navigation is extremely important to the drone fly, as it facilitates the effective exploration and recognition of the unknown environments. Yet, the navigation of drone remains a challenging task, especially for planning the pathway as short as possible to the target/destination whilst avoiding the potential collision with objects in the unexplored space. The conventional navigation heavily relies on the expertise of human, who intuitively designs the drone flyby trajectory based on the spatial layout within the visible range. The resulting navigation system lacks of the globe knowledge of scenes, leading to unsatisfactory or even failed path planning.

To better leverage the global information of 3D environment, researches on drone navigation have focused on collecting and memorizing the environmental information during the navigating process. Typically, the existing works~\cite{henriques2018mapnet,bansal2019combining,bian2019unsupervised} employ the mapping techniques to construct 2D/3D maps with respect to the vacant/occupied space. The mapping result contains rich geometric relationship between objects, which helps to navigate. There have also been navigation approaches based on visual information~\cite{chen2019behavioral, gupta2017cognitive, bansal2019combining}, saving the computational overhead to construct maps. Nonetheless, these works purely condition the accuracy of navigation on either geometric or visual information.

% combine the geometric constrain and visual cue.
% residual updating for reusing the off-the-shelf height map
In this paper, we utilize 2.5D height map for autonomous drone navigation. There are growing computer applications that use height map to represent the boundaries of objects (e.g., buildings or furniture).
Nonetheless, there is nothing guaranteed for the quality of given height maps, as the mapping process likely involves incomplete or out-of-date information.
Here, we advocate the importance of fusing geometric and visual information for a more robust construction of the height map.
The new trend of researches~\cite{tatarchenko2019single, chen2019learning} on the 3D object/scene understanding has also demonstrated that the geometric relationship between objects and visual appearance of scenes are closely correlated. %In our approach, we employ the deep neural network to learn visual and geometric information from the RGB images and height map, respectively. The visual and geometric information are jointly analyzed to extract useful information for the height map refinement. Furthermore, we use the height map along with the sparse 3D keypoints extracted from the sequence of RGB images to construct a set of visual-geometric representations. These representations are applied to update the height map and keypoints, providing accurate information for the following navigation.
We thus propose a \emph{Visual-Geometric Fusion Network} (VGF-Net) to dynamically update the height map during drone navigation by utilizing the timely captured new images (see Figure~\ref{fig:teaser}).

\begin{figure*}[t!]
	\centering
	\includegraphics[width=\linewidth]{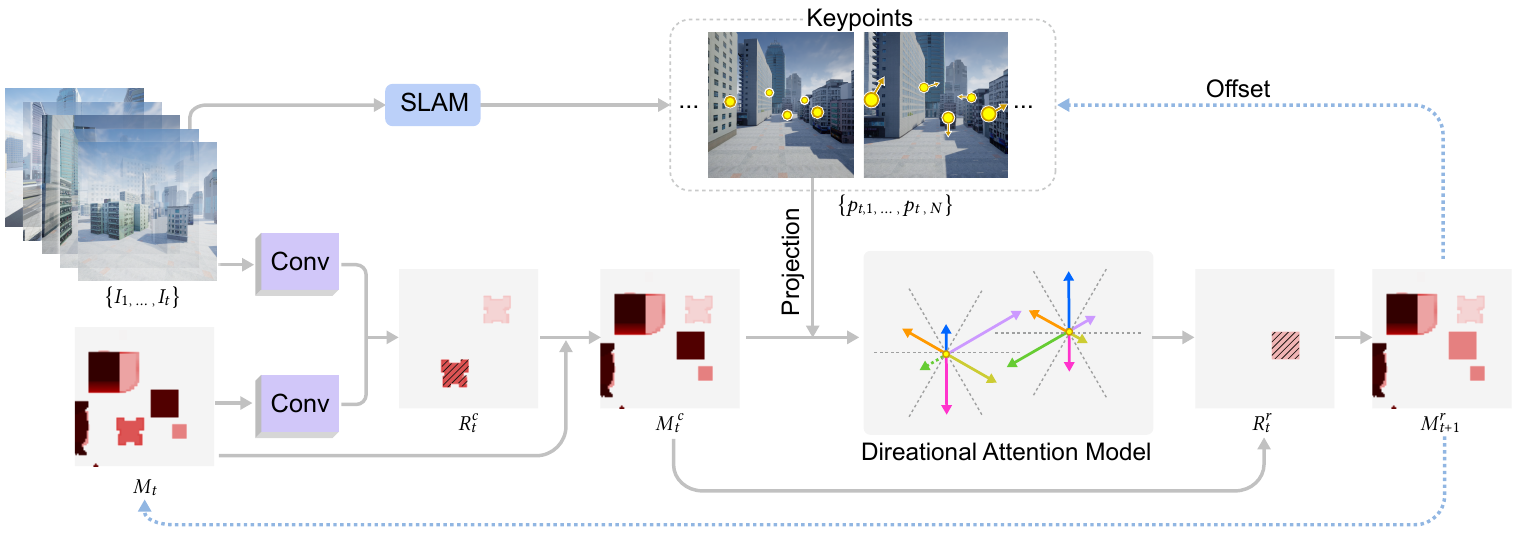}
	\caption{Overview of VGF-Net. At the $t^{th}$ moment, the network uses convolutional layers to learn visual and geometric representations from the RGB image $I_t$ and 2.5D height map $M_t$ (produced at the $(t-1)^{th}$ moment). The representations are combined to compute the residual update map $R^c_t$, which is added to the 2.5D height map to form a renewed height map $M^c_t$. Based on the new height map and the 3D keypoints $\{p_{t,1}, ..., p_{t,N}\}$ (produced by SLAM), we construct the VG representation for each keypoint (yellow dot), which is used by DAM to select useful information to refine object boundaries and 3D keypoints at the next moment. Note that the refined height map $M^r_{t+1}$ is used for path planning, which is omitted for a simple illustration.}
	\label{fig:overview}
\end{figure*}

More specifically, as illustrated in Figure~\ref{fig:overview}, the network takes an initial rough height map together with a sequence of RGB images as input. We use convolutional layers to compute the visual and geometric information to renew the height map. Next, we apply the simultaneous localization and mapping (SLAM)~\cite{mur2017orb} module to extract a sparse set of 3D keypoints from the image sequence. These keypoints are used along with the renewed height map to construct a novel \emph{Visual-Geometric Representation}, which is passed to a \emph{Directional Attention Model}. This attention model exchanges visual and geometric information among objects in the scene, providing quite useful object relationship for simultaneous refinement of the height map and the corresponding keypoints, leading to the successful path planning~\cite{koenig2002d} at each navigation moment. Compared to dense point clouds that require time-consuming depth estimation~\cite{chaurasia2013depth} and costly processing, the sparse keypoints we use are fast to compute yet effective in terms of capturing useful geometric information without much redundancy. As the drone flies over more and more places, our network can achieve and fuse more and more the visual and geometric information to largely increase the precision of height map and consequently the reliability of autonomous navigation.

We intensively train and evaluate our method on a benchmark of seven large-scale urban scenes and six complex indoor scenes for height map construction and drone navigation. The experimental results and comparative statistics clearly demonstrate the effectiveness and the robustness of our proposed VGF-Net.

\section{Related Work}
\label{sec:related}

There have been an array of researches on the navigation system that allows robots to smartly explore the real world. Below, we will mainly survey on the drone navigation and environment mapping, as they are highly relevant to our work in the sense that their navigation systems are driven by the critical environment data.

\subsection{Drone Navigation}

The modern drone systems are generally equipped with various sensors (e.g., RGB-D camera, radar and GPS), which help the hardware devices to achieve accurate perception of the real world. Typically, the data captured by sensors is used for mapping (i.e., the construction of map), providing comprehensive information for planning the moving path of drone. During the navigation process, the traditional methods~\cite{henriques2018mapnet, savinov2018semi} compute the trajectory of drone based on the pre-scribed maps. However, the construction of a precise map is generally expensive and time-consuming. Thus, the recent works~\cite{chen2019behavioral,gupta2017cognitive,bansal2019combining} simplify the construction of map to facility more commercially-cheap navigation.

% without map (visual navigation, other sensors)
The advances on deep learning have significantly improved the robustness of visual navigation, leading to the emergency of many navigation systems that do not rely on the given maps. Kim et al.~\shortcite{kim2015deep} and Padhy et al.~\shortcite{padhy2018deep} use the classification neural network to predict the direction (e.g., right, left or straight) of moving drone. Furthermore, Loquercio et al.~\shortcite{loquercio2018dronet} and Mirowski et al.~\shortcite{mirowski2018learning} use neural networks to compute the angle of flying and the risk of collision, which provide more detailed information to control the drone flyby. Note that the above methods learn the actions of drone from the human annotations. The latest works employ deep reinforcement learning~\cite{tai2017virtual, zhu2017target, wang2019autonomous} to optimize the network, enabling more flexible solutions for autonomous drone navigation in novel environments.

% our map provides visual navigation (cheap and effective) + global recognition of map with geometric constrain
% Compared to the navigation methods w/o map, our method increase the generality power of drone system in different scenes. Compared to the navigation methods w/ map, we provide the accurate update of map
Our approach utilizes  a rough 2.5D height map to increase the success rate of navigation in different complex scenes, which may have various spatial layouts of objects. Compared to the existing methods that conduct the mapping before navigation, we allow for real-time intelligent update of the height map during navigation, largely alleviating negative impacts of problematic mapping results.

\subsection{Mapping Technique}

The mapping technique is fundamental in the drone navigation. The techniques of 2D mapping have been widely used in the navigation task. Henriques et al.~\shortcite{henriques2018mapnet} and Savinov et al.~\shortcite{savinov2018semi} use 2D layout map to store useful information, which is learned by neural networks from the image data of 3D scenes. Chen et al.~\shortcite{chen2019behavioral} use the 2D topological map, which can be constructed using the coarse spatial layout of objects, to navigate the robot in an indoor scene. Different from the methods that consider the 2D map of an entire scene, Gupta et al.~\shortcite{gupta2017cognitive} unify the mapping and 2D path planning to rapidly adjust the navigation with respect to the surrounding local environment. Bansal et al.~\shortcite{bansal2019combining} utilize sparse waypoints to represent the map, which can be used to generate a smooth pathway to the target object or destination.

Compared to 2D mapping, 3D mapping provides much richer spatial information for the navigation system. Wang et al.~\shortcite{wang2017stereo} use visual odometry to capture the geometric relationship between 3D points, which is important to reconstruct the 3D scene. Engel et al.~\shortcite{engel2014lsd, engel2017direct} integrate the tracking of keypoints into the mapping process, harnessing temporal information to produce a more consistent mapping of the global environment. \rev{Futhermore, Huang et al.~\shortcite{huang2020clustervo,ClusterSLAM} use a probabilistic Conditional Random Field model and a noise-aware motion affinity matrix to effectively track both moving and static objects. Wang et al.~\shortcite{piecewiseWang} use plane as a geometric constrain to reconstruct the whole scene.} Besides 3D points, depth information is also important to 3D mapping. During the mapping process, Tateno et al.~\shortcite{tateno2017cnn} and Ma et al.~\shortcite{ma2018sparse} use neural networks to estimate the depth map of a single image, for a faster construction of the 3D map. However, the fidelity of depth estimation is bounded by the scale of training data. To enhance, Kuznietsov et al.~\shortcite{kuznietsov2017semi}, Godard et al.~\shortcite{godard2017unsupervised} and Bian et al.~\shortcite{bian2019unsupervised} train the depth estimation network in semi-supervised/unsupervised manner, where the consistence in-between images are learned.

Nowadays, a vast of real-world 3D models and applications emerge, such as Google earth, and so there is abundant data of height maps available for the training of drone navigation system. Nonetheless, the accuracy and timeliness of such data is impossible to be guaranteed, thus hard to be directly used in practice. We deeply exploit the visual-geometric information fusion representation to effectively and dynamically update the given height map during navigation, yielding a  significant increase of the success rate of the autonomous drone navigation in various novel scenes.

%In this work, we use 2.5D height maps that are easy to obtain or estimate, and . Nowadays a vast of real-world applications have used 2.5D height maps. Thus, there is abundant data of 2.5D height maps for the training of the navigation system. More importantly, we fuse the information of the 2.5D height map and the 3D points, exploiting the visual and 3D geometric information to dynamically update the 2.5D height map for navigation. 
\section{Overview}
\label{sec:overview}

The core idea behind our approach is to fuse the visual and geometric information for the construction of height map. This is done by our \emph{Visual-Geometric Fusion Network} (VGF-Net) to compute the visual-geometric representation with respect to the visual and geometric consistence between the 3D keypoints and object boundaries characterized in the height map. VGF-Net uses the fused representation to refine the keypoints and height map at each moment during drone navigation. Below, we outline the architecture of VGF-Net.

As illustrated in Figure~\ref{fig:overview}, at the $t^{th}$ moment ($t \ge 0$), the network takes the RGB image $I_t$ and the associated height map $M_t$ as input. The image $I_t$ is fed to convolutional layers to compute the visual representation $V_t$. The height map $M_t$ is also input to the convolutional layers for the geometric representation $G_t$. The visual and geometric representations are fused to compute the residual update map $R^c_t$ that updates the height map to $M^c_t$, providing more consistent information for the subsequent steps.

Next, we use the SLAM~\cite{mur2017orb} module to compute a sparse set of 3D keypoints $\{p_{t,1}, ..., p_{t,N}\}$, based on the images $\{I_1, ..., I_t\}$. We project these keypoints to the renewed height map $M^c_t$. For the keypoint $p_{t,i}$, we compute a set of distances $\{d_{t,i,1}, ..., d_{t,i,K}\}$, where $d_{t,i,k}$ denotes the distance from the keypoint $p_{t,i}$ to the nearest object boundary along the $k^{th}$ direction (see Figure~\ref{fig:method}(a)). Intuitively, the keypoint, which is extracted around the objects in the 3D scene, is also near to the boundaries of the corresponding objects in the height map. This relationship between the keypoint $p_{t,i}$ and the object can be represented by the visual and geometric information in the scene. Specifically, this is done by fusing the visual representation $V_t$, geometric representation $G^c_t$ (learned from the renewed height map $M^c_t$) and the distances $\{d_{t,i,1}, ..., d_{t,i,K}\}$ to form a novel \emph{Visual-Geometric} (VG) representation $U_i$ for the keypoint $p_{t,i}$. For all keypoints, we compute a set of VG representations $\{U_{t,1}, ..., U_{t,N}\}$.

Finally, we employ a \emph{Directional Attention Model} (DAM), which takes input as the VG representations $\{U_{t,1}, ..., U_{t,N}\}$, to learn a residual update map $R^r_t$ to refine the height map $M^c_t$. The DAM produces a new height map $M^r_{t+1}$ that respects the importance of each keypoint to the object boundaries in different directions (see Figure~\ref{fig:method}(b)). Meanwhile, we use DAM to compute a set of spatial offsets $\{\Delta p_{t+1,1}, ..., \Delta p_{t+1,N}\}$ to update the keypoints, whose locations are imperfectly estimated by the SLAM. We use the height map $M^r_{t+1}$ for dynamic path planning~\cite{koenig2002d} at the $(t+1)^{th}$ moment, and meanwhile input the image $I_{t+1}$ and the height map $M^r_{t+1}$ to VGF-Net at this moment for next update. As drone flies, the network achieves more accurate information and works more robustly for simultaneous drone navigation and height mapping.

\section{Method}
\label{sec:method}

We now introduce our VGF-Net in more detail. The network extracts visual and geometric information from the RGB images, the associated 2.5D height map and 3D keypoints. In what follows, we formally define the information fusion that produces the visual-geometric representation, which is then used for the refinement of the height map and keypoints.

\subsection{Residual Update Strategy}

The VGF-Net refines the height map and keypoints iteratively, as the drone flies to new places and captures new images. We divide this refinement process into separate moments. At the $t^{th}$ moment, we feed the RGB image \rev{$I_t \in \mathbb{R}^{H_I \times W_I \times 3}$} and the height map\rev{ $M_t \in \mathbb{R}^{H_M \times W_M}$} into the VGF-Net, computing the global visual representation \rev{$V_t \in \mathbb{R}^{H_M \times W_M \times C}$} and the geometric representation \rev{$G_t \in \mathbb{R}^{H_M \times W_M \times C}$} as:
\begin{align}
V_t = \mathcal{F}^{v}(I_t), ~~~G_t = \mathcal{F}^{g}(M_t),
\label{eq:visual_geometric_representation}
\end{align}
where $\mathcal{F}^{v}$ and $\mathcal{F}^{g}$ denote the two sets of convolutional layers. Note that the value of each location on $M_t$ represents the height of object, and we set the height of ground to be 0. We concatenate the representations $V_t$ and $G_t$ for computing a residual update map \rev{$R^c_t \in \mathbb{R}^{H_M \times W_M}$}, which is used to update the height map $M_t$ as:
\begin{align}
M^c_t = M_t+R^c_t,
\label{eq:residual_update_1}
\end{align}
where
\begin{align}
R^c_t = \mathcal{F}^{c}(V_t, G_t).
\label{eq:residual_update_2}
\end{align}
Here, \rev{$M^c_t \in \mathbb{R}^{H_M \times W_M}$} is a renewed height map, and $\mathcal{F}^{c}$ denotes a set of convolutional layers. Compared to directly computing a new height map, the residual update strategy (as formulated by Eq.~\eqref{eq:residual_update_1}) adaptively reuses the information of $M_t$. More importantly, we learn the residual update map $R^c_t$ from the new content captured at the $t^{th}$ moment. It facilitates a more focused update on the height values of regions that are unexplored before the $t^{th}$ moment. The height map $M^c_t$ is fed to an extra set of convolutional layers to produce the representation $G^c_t$, which will be used for the construction of the visual-geometric representation.

\begin{figure}[t!]
	\centering
	\includegraphics[width=\linewidth]{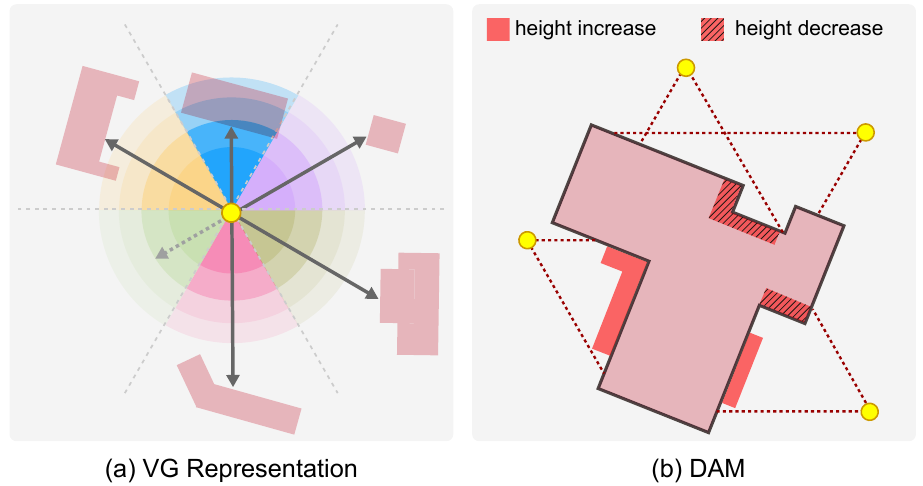}
	\caption{Illustration of fusing visual and geometric information for updating the 2.5D height map. (a) We construct the VG representation for each 3D keypoint (yellow dot) projected to the 2.5D height map. The information of VG representation is propagated to surrounding object boundaries, along different directions (indicated by different colors). The distance between the keypoint and object boundary (black arrow) determines the weight for adjusting the information propagation. The dash arrow means that there is no object along the corresponding direction. (b) Given the existing object boundary, we use DAM to select the most relevant keypoint along each direction. We use the selected keypoints to provide fused visual and geometric information, which is used for refining object boundary. }
	\label{fig:method}
\end{figure}

\subsection{Visual-Geometric Representation}

%\brev{
%We conduct the visual-geometric information fusion to further refine the height map. To capture the geometric relationship between objects, we use a standard SLAM~\cite{mur2017orb} module to extract a sparse set of 3D keypoints $\{p_{t,1}, ..., p_{t,N}\}$ from the sequence of images $\{I_1, ..., I_t\}$. Given the keypoint $p_{t,i} \in \mathbb{R}^{1 \times 3}$, we project it to the 2.5D space as:
%\begin{align}
%p^{\prime}_{t,i} = p_{t,i}RS + T.
%\label{eq:projection_1}
%\end{align}
%Here, $T \in \mathbb{R}^{1 \times 3}$, $R \in \mathbb{R}^{3 \times 3}$, and $S \in \mathbb{R}^{3 \times 3}$ represent the translation, rotation and scaling matrices, respectively, which in our framework are defaulted to be as:
%\begin{align}
%T \!=\! \begin{bmatrix} 336 \!&\! 0 \!&\! -336 \end{bmatrix}, R \!=\! \begin{bmatrix} 0 \!&\! 0 \!&\! 1 \\ 0 \!&\! 1 \!&\! 0 \\ -1 \!&\! 0 \!&\! 0 \end{bmatrix}, S \!=\! \begin{bmatrix} 1 \!&\! 0 \!&\! 0 \\ 0 \!&\! 1 \!&\! 0 \\ 0 \!&\! 0 \!&\! 1 \end{bmatrix}.
%\label{eq:projection_2}
%\end{align}
%}

We conduct the visual-geometric information fusion to further refine the height map. To capture the geometric relationship between objects, we use a standard SLAM~\cite{mur2017orb} module to extract a sparse set of 3D keypoints $\{p_{t,1}, ..., p_{t,N}\}$ from the sequence of images $\{I_1, ..., I_t\}$. Given the keypoint $p_{t,i} \in \mathbb{R}^{1 \times 3}$ in the camera coordinate system, we project it to the 2.5D space as:
\begin{align}
p^{\prime}_{t,i} = p_{t,i}SR + T.
\label{eq:projection_1}
\end{align}
Here, $S \in \mathbb{R}^{3 \times 3}$ is decided by a pre-defined scale factor, which could be calculated at the initialization of the SLAM system or by GPS adjustment. $T \in \mathbb{R}^{1 \times 3}$ and $R \in \mathbb{R}^{3 \times 3}$ translate the origin of the 3D point set from the camera to the height map coordinate system. In the height map coordinate system, the drone is located at\rev{ $(\frac{W}{2},0)$}, where $W$ represent the \rev{width} of the height map.
\label{eq:projection_2}

Note that the first two dimensions of $p^{\prime}_{t,i} \in \mathbb{R}^{1 \times 3}$ indicate the location on the height map, and the third dimension indicates the corresponding height value. The set of keypoints $\{p^{\prime}_{t,1},...,p^{\prime}_{t,N}\}$ are used for constructing the visual-geometric representations.

Next, for each keypoint $p^{\prime}_{t,i}$, we compute its distances to the nearest objects in $K$ different directions. Here, we refer to objects as the regions that have larger height values than the ground (with height value of 0) in the height map $M^c_t$. As illustrated in Figure~\ref{fig:method}(a), we compute the Euclidean distance $d_{t,i,k}$ along the $k^{th}$ direction, from $p^{\prime}_{t,i}$ to the first location, where the height value is larger than 0. We compute a set of distances $\{d_{t,i,1}, ..., d_{t,i,K}\}$ for $K$ directions, then use $V_t$ (see Eq.~\eqref{eq:visual_geometric_representation}), $G^c_t$ and this distance set to form the VG representation $U_{t,i} \in \mathbb{R}^{K}$ as:
\begin{align}
U_{t,i,k} = \mathcal{F}^v_k(W_{t,i,k} V_t) + \mathcal{F}^g_{k}(W_{t,i,k} G^c_{t,i}),
\label{eq:VG_1}
\end{align}
where
\begin{align}
W^v_{t,i,k} = \sum^{K}_{k^\prime=1} \exp(-|d_{t,i,k}-d_{t,i,k^\prime}|).
\label{eq:VG_2}
\end{align}
Here, $G^c_{t,i} \in \mathbb{R}^{C}$ denotes the feature vectors located in $p^{\prime}_{t,i}$ in the map $G^c_t$. In Eq.~\eqref{eq:VG_1},  $U_{t,i,k}$ is represented as a weighted map with the resolution equal to the geometric representation ($20 \times 20$ by default), where $W_{t,i,k}$ plays as a weight of importance that is determined by the distance from the keypoint $p^{\prime}_{t,i}$ to the nearest object boundary along the $k^{th}$ direction. As formulated in Eq.~\eqref{eq:VG_1} and Eq.~\eqref{eq:VG_2}, longer distance decays the importance. Besides, we use independent set of fully connected layers (i.e., $\mathcal{F}^v_k$ and $\mathcal{F}^g_{k}$ in Eq.~\eqref{eq:VG_1}) to learn important information from $V_t$ and $G^c_{t,i}$. It allows the content, which is far from $p^{\prime}_{t,i}$, to have the opportunity to make an impact on $U_{t,i,k}$. We construct the VG representation for each keypoint in $\{p^{\prime}_{t,1},...,p^{\prime}_{t,N}\}$, while each VG representation captures the visual and geometric information around the corresponding keypoint. Based on the the VG representations, we propagate the information of the keypoints to each location on the height map, where the corresponding height value is refined. We also learn temporal information from the VG representations to refine the spatial locations of keypoints at the $(t+1)^{th}$ moment, as detailed below.

\begin{figure*}[t!]
	\centering
	\includegraphics[width=\linewidth]{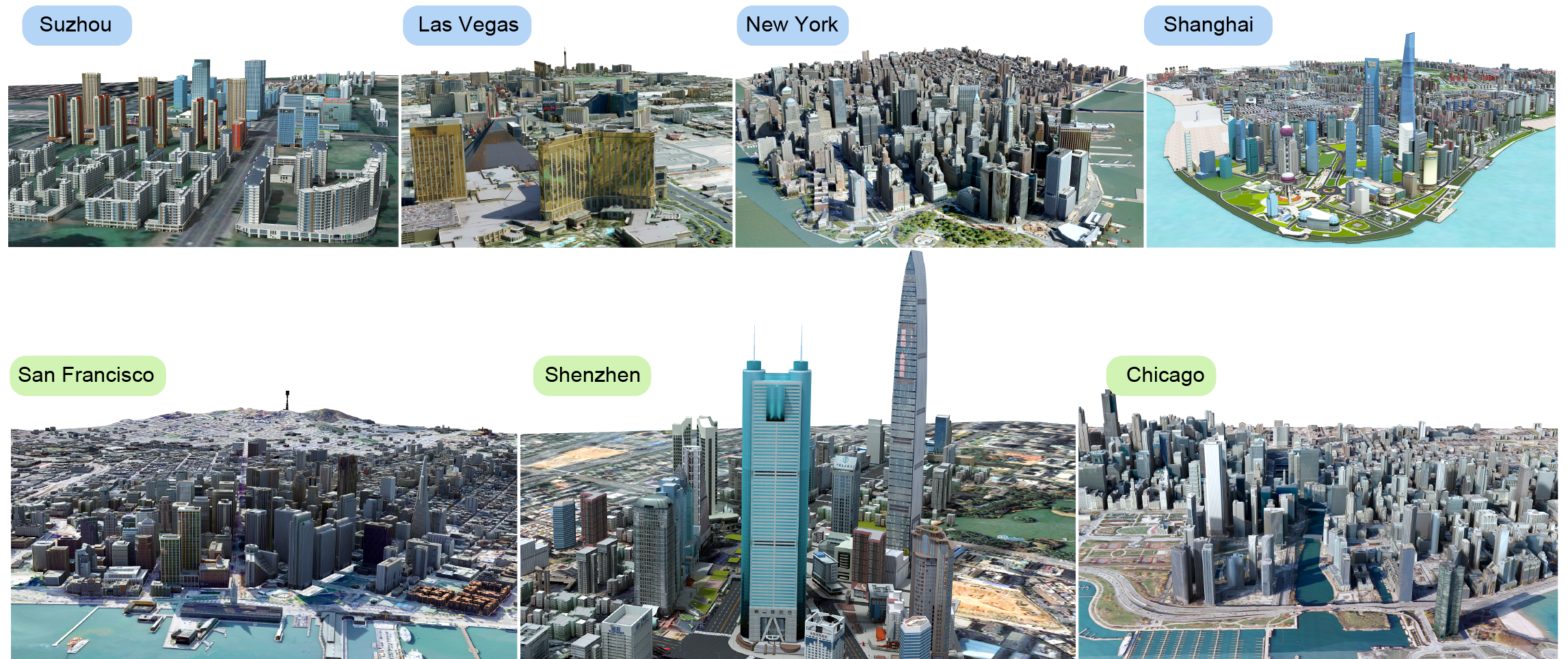}
	\caption{Overview of our 3D urban navigation dataset, including 7 city scenes with different characteristics.}
	\label{fig:dataset}
\end{figure*}

\subsection{Directional Attention Model}

% refine the 2.5D height map
We use DAM to propagate the visual and geometric information, from each keypoint to each location on the height map, along different directions. More formally, for a location $p^h_j \in \mathbb{R}^{1 \times 3}$ on the height map $M^c_t$, we conduct the information propagation that yields a new representation $Q_{t,j} \in \mathbb{R}^{C \times K}$ as:
\begin{align}
Q_{t,j} = \sum^{N}_{i=1} G^c_{t,j} U_{t,i}^\top.
\label{eq:DAM_1}
\end{align}
Along the second dimension of the representation $Q_{t,j}$, we perform max pooling to yield $Q^{\prime}_{t,j} \in \mathbb{R}^{C}$ as:
\begin{align}
Q^{\prime}_{t,j,c} = \max(Q_{t,j,c,1},...,Q_{t,j,c,K}).
\label{eq:DAM_2}
\end{align}
As illustrated in Eq.~\eqref{eq:DAM_1}, $Q_{t,j,c,k}$ summarizes the influence of all keypoints along $k^{th}$ direction. We perform max pooling on the set $\{Q_{t,j,c,1},...,Q_{t,j,c,K}\}$ (see Eq.~\eqref{eq:DAM_2}), attending to the most information along a direction to form the representation $Q^{\prime}_{t,j,c}$ (see Figure~\ref{fig:method}(b)). To further refine the height map, we use the representation \rev{$Q^{\prime}_{t} \in \mathbb{R}^{H_M \times W_M \times C}$} to compute another residual update map \rev{$R^r_t \in \mathbb{R}^{H_M \times W_M}$}, which is added to the height map $M^c_t$ to form a new height map \rev{$M^r_{t+1} \in \mathbb{R}^{H_M \times W_M}$} as:
\begin{align}
M^r_{t+1} = M^c_t+R^r_t,
\label{eq:DAM_3}
\end{align}
where
\begin{align}
R^r_t = \mathcal{F}^{r}(V_t, Q^{\prime}_{t}).
\label{eq:DAM_4}
\end{align}
Again, $\mathcal{F}^{r}$ denotes a set of convolutional layers. We make use of the new height map $M^r_{t+1}$ for the path planning at the $(t+1)^{th}$ moment.

% refine the 3D keypoints
We refine not only the 2.5D height map but also the 3D keypoints at the $(t+1)^{th}$ moment. Assume that we use SLAM to produce a new set of keypoints $\{p_{t+1,1},...,p_{t+1,N}\}$. We remark that the keypoint sets at the $t^{th}$ and $(t+1)^{th}$ moments are not necessary the same. To refine the new keypoint $p_{t+1,j} \in \mathbb{R}^{1 \times 3}$, we use DAM to compute the representation $\Delta p^{\prime}_{t+1,j} \in \mathbb{R}^{3 \times K}$ as:
\begin{align}
\Delta p^{\prime}_{t+1,j} = \sum^{N}_{i=1} p_{t,i} U_{t,i}^\top.
\label{eq:DAM_5}
\end{align}
In this way, DAM distills the information of keypoints at the $t^{th}$ moment, which is propagated to the next moment. Again, we use max pooling to form the spatial offset $\Delta p_{t+1,j,c} \in \mathbb{R}^{1 \times 3}$ for updating keypoint $p_{t+1,j}$ as:
\begin{align}
\Delta p_{t+1,j,c} = \max(\Delta p^{\prime}_{t+1,j,c,1},...,\Delta p^{\prime}_{t+1,j,c,K}).
\label{eq:DAM_6}
\end{align}
We take the average of the updated keypoints $p_{t+1,j}+\Delta p_{t+1,j}$ and the estimated keypoints $p_{t+1,j}$ in place of the original one to construct the VG representation at the $(t+1)^{th}$ moment.

\subsection{Training Details}

We use the $L_1$ loss function for training the VGF-Net as:
\begin{align}
\mathcal{L}(M^{gt}_t, M^r_t) = \sum^{T}_{t=1} \sum^{H \times W}_{j=1} |M^{gt}_{t, j} - M^r_{t, j}|,
\label{eq:loss_function}
\end{align}
where $M^{gt}_t \mathbb{R}^{H \times W}$ is the ground-truth height map. Actually, we select 8 pairs of RGB image and height map ($T=8$) to construct each mini-batch for the standard SGD solver. We set the height and width of each RGB image ($224 \times 224$) and the height map ($20 \times 20$). The overall training samples is nearly 24000 images randomly sampled in 3 scenes, while we test the model on the 24000 samples sampled on the other 3 scenes. Details about the dataset could be found in Sec.~\ref{sec:results}. We train the network for 30 epochs, and use the final snapshot of network parameters for testing. The learning rate is set to 0.001 at the first 15 epochs, and decayed to 0.0001 for a more stable optimization.

By default, the backbone of $\mathcal{F}^{v}$ and $\mathcal{F}^{g}$ is a ResNet-18, while the remained $\mathcal{F}^{c}$ and $\mathcal{F}^{r}$ is two stacked $3 \times 3$ convolutional layer with max-pooling and batch normalization.

Note that it is our contribution to learn spatial offsets of 3D keypoints, without explicitly using any ground-truth data. This is done by modeling the computation of spatial offsets as a differentiable function with respect to the VG representation. In this way, we enable the end-to-end learning of spatial offsets, where the related network parameters can be optimized by the back-propagated gradients. It significantly reduces the effort for data annotation, while allows the network training to be flexibly driven by data.

When constructing the VG representation, we set the number of directions $K=16$ for each keypoint, and the number of keypoints $N=50$ at each moment. We remark that these hyper-parameters are chosen based on the validation results.

\section{Results and Discussion}
\label{sec:results}

\setlength{\tabcolsep}{5pt}
\renewcommand{\arraystretch}{1.5}
\begin{table*}[!t]
	\caption{Statistics of our 3D urban navigation dataset. Note that in addition to buildings, there may also exist many other objects we must consider, such as trees, flower beds, and street lights, which highly increase the challenge for height mapping and autonomous navigation task.}
	\label{tab:dataset}
	%\begin{minipage}{\columnwidth}
	\begin{center}
	\begin{tabular}{c||c|c|c|c|c}
		\hline
		  scene     & area ($km^2$)& objects (\#) & model size ($MB$) & \rev{texture images} (\#) & texture size ($MB$) \\ \hline
		 New York    & 7.4 & 744  &86.4 & 762 & 122 \\ \hline
		 Chicago & 24 & 1629  &146 & 2277 & 227 \\ \hline
	     San Francisco  & 55 & 2801 & 225 & 2865 & 322 \\ \hline
         Las Vegas  & 20 & 1408  & 108 & 1756 & 190 \\ \hline
         Shenzhen  & 3 & 1126   & 50.3 & 199 & 72.5 \\ \hline
         Suzhou  & 7 & 168  & 191& 395 & 23.7\\ \hline
         Shanghai  & 37 & 6850   & 308 & 2285 & 220 \\ \hline
	\end{tabular}
	\end{center}
	%\end{minipage}
\end{table*}

\setlength{\tabcolsep}{5pt}
\renewcommand{\arraystretch}{1.5}
\begin{table*}[!t]
	\caption{Comparisons with different strategies of information fusion, in terms of the accuracy of height mapping (average $L_1$ error). We also show the accuracies ($\%$) of predicting height values, with respect to different ranges of error ($<$ 3$m$, 5$m$ and 10$m$). All strategies are evaluated on the testing (i.e., unknown and novel) scenes of San Francisco, Shenzhen and Chicago. }
	\label{tab:result}
	%\begin{minipage}{\columnwidth}
	\begin{center}
		\begin{tabular}{c||c|c|c||c|c|c}
			\multirow{2}{*}{method}        & \multicolumn{3}{c||}{average $L_1$ error ($m$)}            & \multicolumn{3}{c}{ accuracy w.r.t. error $\in [0,3]m$ (\%) }                                                                 \\ \cline{2-7}
			                               & San Francisco                                              & Shenzhen                                                  & Chicago       & San Francisco & Shenzhen      & Chicago       \\ \hline
			w/o fusion                     & 4.57                                                       & 4.57                                                      & 4.49          & 68.95\%       & 68.02\%       & 70.05\%       \\ \hline
			w/ fusion                      & 2.37                                                       & 2.93                                                      & 3.41          & 85.09\%       & 83.63\%       & 78.44\%       \\ \hline
			w/ fusion and memory           & 2.81                                                       & 3.44                                                      & 4.02          & 79.86\%       & 79.20\%       & 72.86\%       \\ \hline
			w/ fusion, memory and exchange & 2.35                                                       & 3.04                                                      & 3.80          & 80.54\%       & 82.36\%       & 74.73\%       \\ \hline
			full strategy                  & {\bf 1.98}                                                 & {\bf 2.72}                                                & {\bf 3.10}    & {\bf 85.71\%} & {\bf 86.13\%} & {\bf 80.46\%} \\ \hline\hline
			\multirow{2}{*}{method}        & \multicolumn{3}{c||}{accuracy w.r.t. error $\in [0,5]m$ (\%) } & \multicolumn{3}{c}{accuracy w.r.t. error $\in [0,10]m$ (\%)}                                                                  \\ \cline{2-7}
			                               & San Francisco                                              & Shenzhen                                                  & Chicago       & San Francisco & Shenzhen      & Chicago       \\ \hline
			w/o fusion                     & 75.02\%                                                    & 74.08\%                                                   & 76.86\%       & 83.96\%       & 83.96\%       & 85.71\%       \\  \hline
			w/ fusion                      & 89.20\%                                                    & 87.39\%                                                   & 84.12\%       & 93.87\%       & 92.25\%       & 91.18\%       \\  \hline
			w/ fusion and memory           & 86.35\%                                                    & 84.56\%                                                   & 80.36\%       & 93.00\%       & 91.31\%       & 89.51\%       \\  \hline
			w/ fusion, memory and exchange & 86.13\%                                                    & 86.43\%                                                   & 81.41\%       & 93.33\%       & 91.85\%       & 89.94\%       \\  \hline
			full strategy                  & {\bf 89.22\%}                                              & {\bf 88.90\%}                                             & {\bf 85.30\%} & {\bf 94.10\%} & {\bf 92.56\%} & {\bf 91.67\%} \\  \hline
		\end{tabular}
	\end{center}
	%\end{minipage}
\end{table*}

\begin{figure}[t!]
	\centering
	\includegraphics[width=\linewidth]{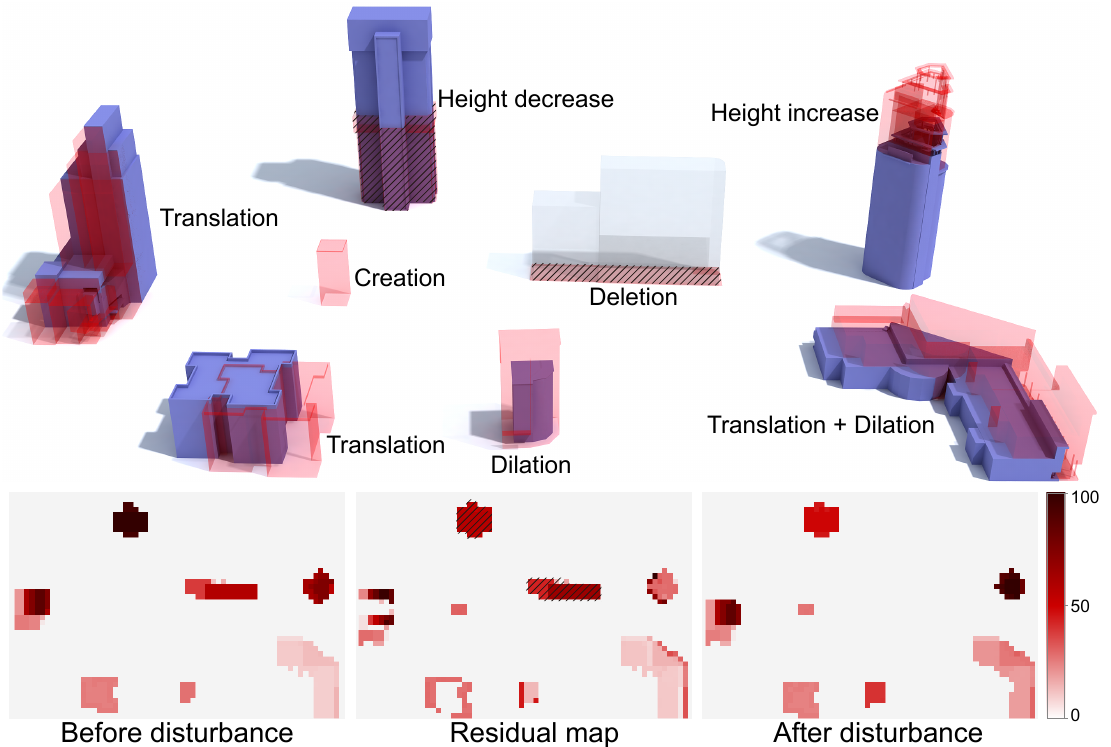}
	\caption{Illustration of disturbance manipulations. Actually, these manipulations can be combined to yield the disturbance results (e.g., translation and dilation). The bottom row of this figure shows the difference between height maps before/after disturbance. The residual map is learned by our VGF-Net, for recovering the disturbed height map to the undisturbed counterpart.}
	\label{fig:disturbance}
\end{figure}

\subsection{Description of Experimental Dataset}

To promote the related research on drone navigation, we newly collect a 3D urban navigation dataset. This dataset contains 7 models of different city scenes (see Figure~\ref{fig:dataset}).

Note that New York, Chicago, San Francisco, and Las Vegas are Google Earth models we download, which are similar to the real-world scenes with respect to the appearance but most objects inside are only buildings. We have also Shenzhen, Suzhou and Shanghai that are manually built based on the map by professional modelers, which contain rich 3D objects (e.g., buildings, trees, street lights and road signs, etc.) and other stuff (e.g., ground, sky and sea). There are various spatial configurations of objects, building styles and weather conditions in these 3D scenes. Thus, we provide challenging data for evaluating the navigation system.
The models are input to the render for producing sequences of RGB images. All RGB images and the associated 2.5D height maps are used to form a training set (i.e., New York, Las Vegas and Suzhou) and a testing set (i.e., San Francisco, Shenzhen, and Chicago). We provides more detailed statistics of the dataset in Table~\ref{tab:dataset}.

%Note that all 3D models are synthesized, as illustrated in Fig.~\ref{fig:dataset}. To achieve these models, we use Google Earth to provide the ground-truth 2.5D height map, determining the locations object and stuff. Based on the 2.5D height map, we use 3D Max to build 3D models. Additionally, we choose suitable RGB images as the texture of 3D models, increasing visual details of 3D models and therefore providing data similar to the real-world scene.

To train our VGF-Net, which takes as input a rough imperfect height map and outputs an accurate height map, we use 5 types of manipulations (i.e., translation, height increase/decrease, size dilation/contraction, creation and deletion) to disturb the object boundaries in the ground-truth height map. One time of the disturbance increases or decreases height values by 10$m$ in certain map locations. See Figure~\ref{fig:disturbance} for an illustration of our manipulations.

\begin{figure*}[t!]
	\centering
	\includegraphics[width=\linewidth]{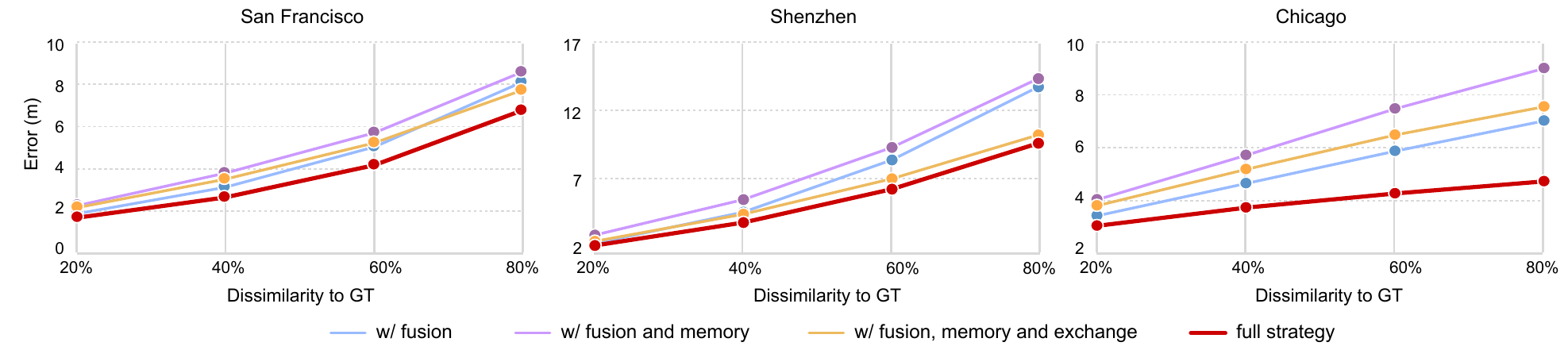}
	\caption{We disturb the 2.5D height maps, which are used to examine the robustness of different information fusion approaches. We evaluate different approaches on the testing sets of San Francisco, Shenzhen and Chicago. All results are reported in terms of $L_1$ errors.}
	\label{fig:stress}
\end{figure*}

\subsection{Different Strategies of Information Fusion}
\label{subsec:internal}

The residual update, VG representation and DAM are critical components of VFG-Net, defining the strategy of information fusion. Below, we conduct an internal study by removing these components, and examine the effect on the accuracy of height mapping (see Table~\ref{tab:result}).

First, we report the performance using visual information only for height mapping, disabling any visual and geometric fusion. Here, the visual information is learned from RGB images (see the entries ``w/o fusion" in Table~\ref{tab:result}). But visual information is insufficient for reconstructing height maps, which requires the modeling of geometric relationship between objects, yielding lower performances compared to other methods using geometric information.

Next, we examine the efficiency of residual update strategy. At each moment, the residual update allows VGF-Net to reuse the mapping result produced earlier. This strategy, where the useful visual and geometric contents can be effectively distilled and memorized at all moments, improves the reliability of height mapping. Thus, by removing the residual update (see the entries ``w/ fusion" in Table~\ref{tab:result}) from VGF-Net (see the entries ``full strategy"), we degrade the performance of height mapping.

We further study the effect of VG representation on the performance. The VG representation can be regarded as an information linkage. It contains fused visual and geometric information, which is exchanged among objects. Without the VG representation, we use independent sets of convolutional layers to extract the visual and geometric representations from the image and height map, respectively. The representations are simply concatenated for computing the residual update map (see the entries ``w/ fusion and memory" in Table~\ref{tab:result}). This manner successfully disconnects the communication between objects and leads to performance drops on almost all scenes, compared to our full strategy of information fusion.

We find that the performance of using memory of height values lags behind the second method without using memory (see the entries ``w/ fusion"  in Table~\ref{tab:result}). We explain that the information fusion with memory easily accumulates errors in the height map over time. Thus, it is critical to compute the VG representation based on the memorized information, enabling the information exchange between objects (see the entries ``w/ fusion, memory and exchange"). Such exchange process provides richer object relationship to effectively address the error accumulation problem, significantly assisting height mapping at each moment.

Finally, we investigate the importance of DAM (see the entries ``w/ fusion, memory and exchange" in Table~\ref{tab:result}). We solely remove DAM from the full model, by directly using VG representations to compute the residual update map and spatial offsets for refining the height map and keypoints. Compared to this fusion strategy, our full strategy with DAM provides a more effective way to adjust the impact of each keypoint along different directions. Therefore, our method achieves the best results on all testing scenes.

\subsection{Sensitivity to the Quality of Height Map}

\begin{figure*}[t!]
	\centering
	\includegraphics[width=\linewidth]{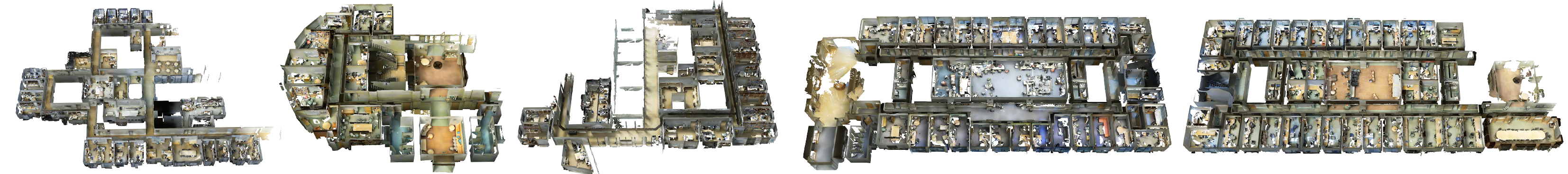}
	\caption{The five indoor training scenes selected from the S3DIS dataset\protect~\cite{armeni20163d}.}
	\label{fig:indoor_train}
\end{figure*}

\begin{figure*}[t!]
	\centering
	\includegraphics[width=\linewidth]{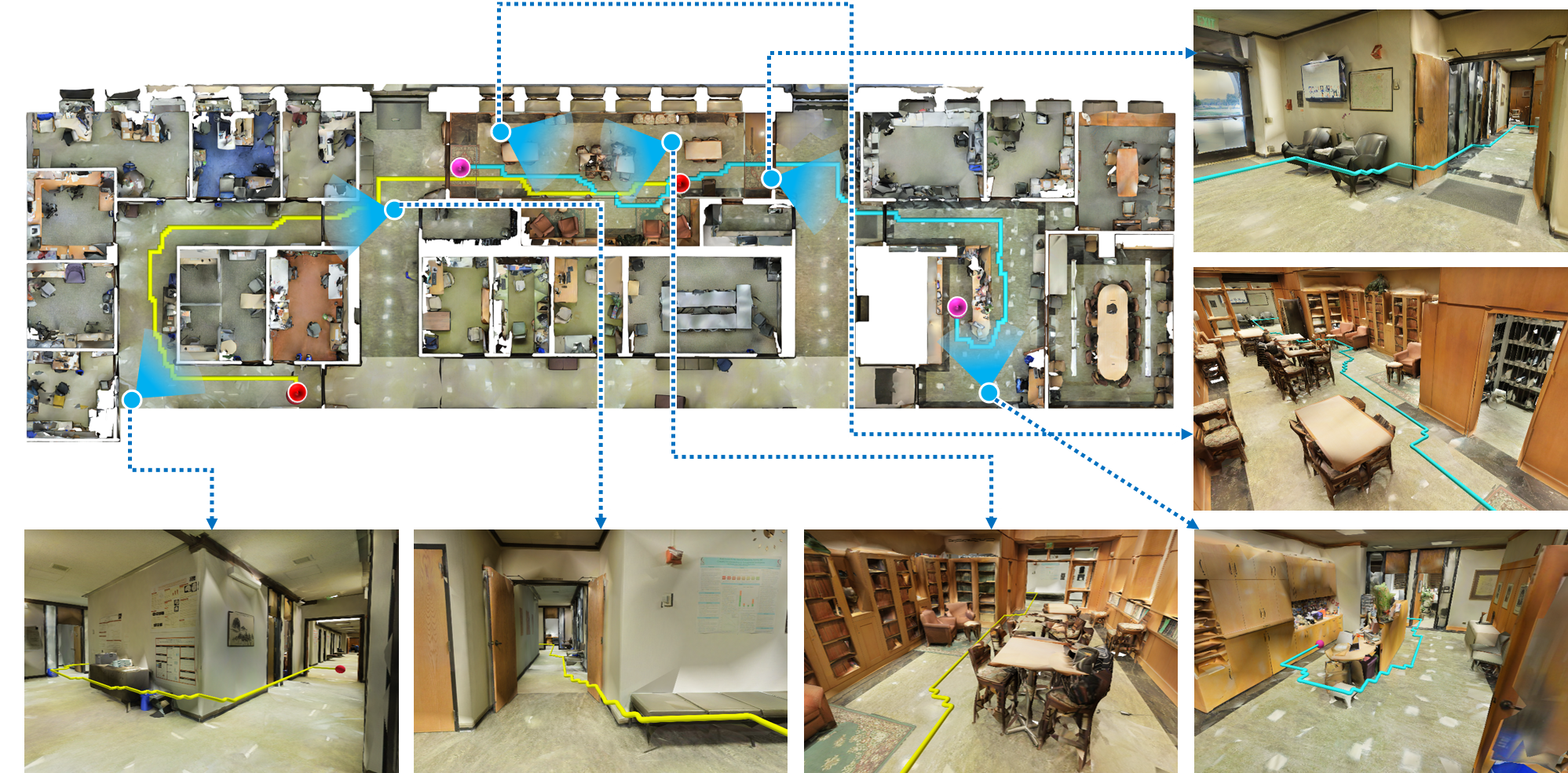}
	\caption{The successful navigation trajectories produced by VGF-Net in a complicate indoor testing scene from the S3DIS dataset\protect~\cite{armeni20163d}.}
	\label{fig:indoor_test}
\end{figure*}

\setlength{\tabcolsep}{5pt}
\renewcommand{\arraystretch}{1.5}
\begin{table*}[!t]
	\caption{We compare VGF-Net with/without using depth to other methods. All methods are evaluated on the outdoor sets (i.e., San Francisco, Shenzhen and Chicago) and the indoor set (i.e., S3DIS). Results are reported in terms of the success rates of navigation.}
	\label{tab:navi}
	%\begin{minipage}{\columnwidth}
	\begin{center}
	\begin{tabular}{c||c|c|c||c|c|c}
		\multirow{2}{*}{outdoor test} & \multicolumn{3}{c||}{w/ depth}    & \multicolumn{3}{c}{w/o depth}        \\ \cline{2-7}
								 &   ground-truth depth        & \multicolumn{2}{c||}{estimated depth~\cite{bian2019unsupervised}} & \multicolumn{3}{c}{VGF-Net}                   \\ \cline{1-7}
								 San Francisco             & 100\%         & \multicolumn{2}{c||}{27\%}          & \multicolumn{3}{c}{{\bf 85\%}}               \\ \cline{1-7}
								 Shenzhen                  & 100\%         & \multicolumn{2}{c||}{34\%}          & \multicolumn{3}{c}{{\bf 83\%}}               \\ \cline{1-7}
								 Chicago                   & 100\%         & \multicolumn{2}{c||}{19\%}          & \multicolumn{3}{c}{{\bf 82\%}}               \\ \hline \hline
		\multirow{2}{*}{indoor test}  & \multicolumn{3}{c||}{w/ depth}    & \multicolumn{3}{c}{w/o depth}         \\ \cline{2-7}
			&  LSTM~\cite{gupta2017cognitive}  & CMP~\cite{gupta2017cognitive}  & VGF-Net & LSTM~\cite{gupta2017cognitive}  & CMP~\cite{gupta2017cognitive}  & VGF-Net        \\ \cline{1-7}
								 S3DIS                     & 71.8\%          & 78.3\%   &   {\bf 92\%}     & 53\%        & 62.5\%       & {\bf 76\%}     \\ \hline
		\end{tabular}
	\end{center}
	%\end{minipage}
\end{table*}

%As demonstrated in the above experiment, it is important to the iterative information fusion for achieving a more global understanding of 3D scene to perfect the 2.5D height map. During the iterative procedure, the problematic height values may be memorized to make a negative impact on the production of height map at future moment. In this experiment, we investigate the sensitivity of different approaches to the quality of height map, by controlling the percentage of height values that are dissimilar to the ground-truth heights. We produce these dissimilar height values by randomly using 5 types of manipulations (i.e., translation, height changing, size dilation, creation and deletion) to disturb the object boundaries in the height map. The disturbance increases or decreases the height values by 10$m$, in some locations of the height map. See the illustration in Fig.~\ref{fig:disturbance} for more details of our manipulations.

As demonstrated in the above experiment, it is important to the iterative information fusion for achieving a more global understanding of 3D scene to perfect the height map estimation. During the iterative procedure, the problematic height values may be memorized to make a negative impact on the production of height map at future moment. In this experiment, we investigate the sensitivity of different approaches to the quality of height maps, by controlling the percentage of height values that are dissimilar to the ground-truth height maps. Again, we produce dissimilar height maps by using disturbance manipulations to change the object boundaries.

At each moment, the disturbed height map is input to the trained model to compute the new height map, which is compared to the ground-truth height map for calculating the average $L_1$ error. In Figure~\ref{fig:stress}, we compare the average $L_1$ errors produced by 4 different information fusion strategies (i.e., see the entries ``w/ fusion", ``w/ fusion and memory", ``w/ fusion, memory and exchange" and ``full strategies" in Table~\ref{tab:result}), which learn geometric information from height maps. As we can see, heavier disturbances generally lead to the degradation of all strategies.

The strategy ``w/ fusion and memory" performs the worst among all approaches, showing very high sensitivity to the quality of height maps. This result further evidences our finding in Sec.~\ref{subsec:internal}, where we have shown the unreliability of the method with memory of height information but without information exchange. Compared to other methods, our full strategy yields better results. Especially, given a very high percentage (80\%) of incorrect height values, our full strategy outperforms other methods by remarkable margins. These results clearly demonstrate the robustness of our strategy.

\subsection{Comparison on the Navigation Task}

The quality of 2.5D height maps, which are estimated by the height mapping, largely determines the accuracy of drone navigation. In this experiment, we compare our VGF-Net to different mapping approaches. All methods are divided into two groups. In the first group, the approaches apply depth information for height mapping. Note that the depth information can be achieved by scanner~\cite{gupta2017cognitive}, or estimated by deep network based on the RGB images~\cite{bian2019unsupervised}. The second group consists of approaches that only use RGB images to reconstruct the height map. In addition to an initial height map that can be easily obtained from various resources, our VGF-Net only requires image inputs, but can also accept depth information if available without changing any scheme architecture. We set the height of flight to be 10$\sim$30$m$ for drone, evaluating the success rate of 3D navigation on our outdoor dataset. Overheight (e.g., 100$m$) always leads to successful navigation, making the evaluation meaningless. On the indoor dataset~\cite{armeni20163d} (see also Figure~\ref{fig:indoor_train} and Figure~\ref{fig:indoor_test}) , we report the success rate of 2D drone navigation, by fixing the height of flight to 0.5$m$. All results can be found in Table~\ref{tab:navi}.

\begin{figure}[t!]
	\centering
	\includegraphics[width=\linewidth]{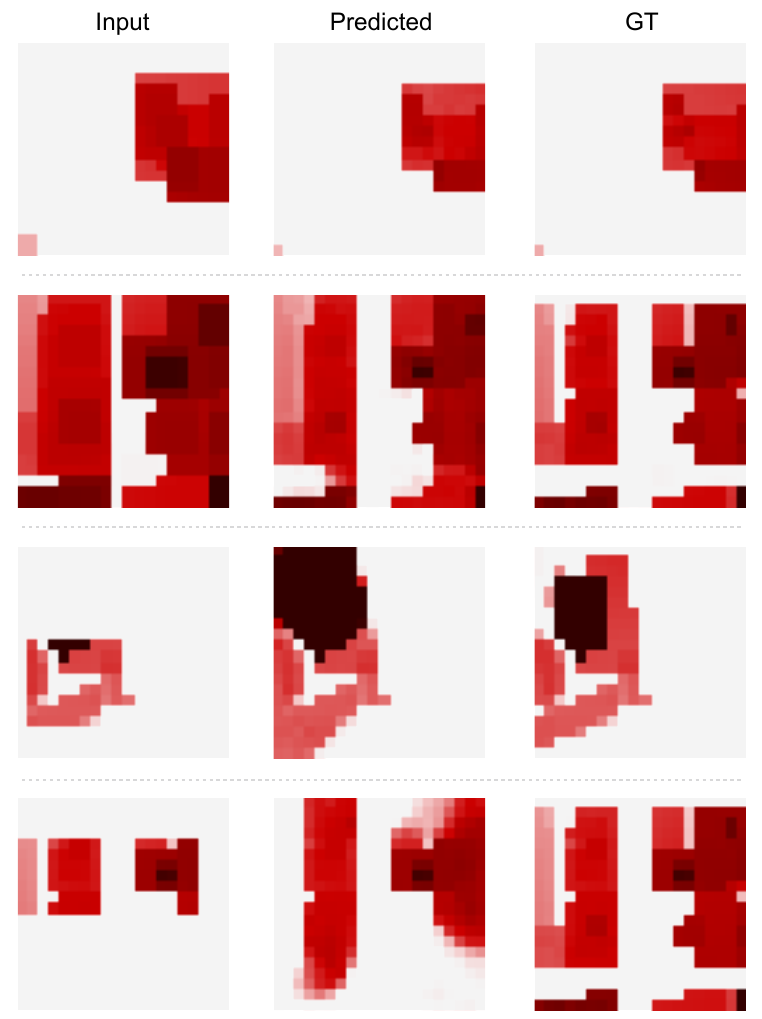}
	\caption{Examples of height mapping. All the height maps are selected from the outdoor dataset. Here, we compare the height maps with noise (in the first column), predicted height maps (in the second column) and ground-truth height maps (in the last column).}
	\label{fig:height_mapping_results}
\end{figure}

%We evaluate different approaches on the outdoor dataset.
Obviously, using accurate depth information can yield a perfect success rate of navigation (see the entry ``ground-truth depth"). Here, the depth data is directly computed from the synthesized 3D urban scenes, without involving any noise. However, due to the limitation of hardware device, it is difficult for the scanner to really capture the accurate depth data of outdoor scenes. A simple alternative is to use deep network to estimate the depth based on the RGB image (see the entry ``estimated depth"). Depth estimation often produces erroneous depth values for the height mapping, even with the most advanced method~\cite{bian2019unsupervised}, thus severely misleading the navigation process. Similar to depth information, the sparse 3D keypoints used in our approach also provide valuable geometry information of objects. More importantly, our VGF-Net uses visual cues to assist the learning of geometric representations. Therefore, our approach without using depth produces better results than that of using depth estimated by state-of-the-art techniques. We have shown an example of trajectory for 3D drone navigation in Figure~\ref{fig:teaser}. We also show examples of height mapping in Figure~\ref{fig:height_mapping_results}, where the height map with redundant boundary (see the first two rows of Figure~\ref{fig:height_mapping_results}) or missing boundary (see the last two rows of Figure~\ref{fig:height_mapping_results}) is input to the VGF-Net. Even given the input height maps with much noise, our network still precisely recovers the height information.

Depth data of indoor scenes (see Figure~\ref{fig:indoor_train}) can be more easily achieved. With the available depth information, we can trivially input the RGB image along with the associated depth to the VGF-Net, producing the height map. We compare VGF-Net to the recent approach~\cite{gupta2017cognitive} (see the entries ``LSTM " and ``CMP") that produces state-of-the-art indoor navigation accuracies. Our method achieves a better result under the same condition of training and testing. Without depth, our approach still leads to the best result among all image based methods. It demonstrates the generality and ability of our approach, in terms of stably learning useful information from different data sources. In Figure~\ref{fig:indoor_test}, we show more navigation trajectories planned by our approach in an indoor testing scene.

\section{Conclusions and Future Work}
\label{sec:conclusion}

The latest progress on drone navigation is largely driven by the active sensing and selecting the useful visual and geometric information of surrounding 3D scenes. In this paper, we have presented VGF-Net, where we fuse visual and geometric information for simultaneous drone navigation and height mapping. Our network distills the fused information, which is learned from the RGB image sequences and an initial rough height map, constructing a novel VG representation to better capture object/scene relation information. Based on the VG representation, we propose DAM to establish information exchange among objects and select essential object relationship in a data-driven fashion. By using residual update strategy, DAM progressively refines the object boundaries in the 2.5D height map and the extracted 3D keypoints, showing its generality to various complicate outdoor/indoor scenes. \rev{The mapping module runs at nearly 0.2sec on a mobile GPU, which could be further optimized by compression and pruning in an embedded system.}

VGF-Net eventually outputs the residual update map and spatial offsets, which are used for explicitly updating the geometric information of objects (i.e., the 2.5D height map and 3D keypoints). It should be noted that we currently use convolutional layers to learn implicit representation from the fused information, and update the visual representation. The visual content of the sequence of RGB image shows complex patterns, which together form the global object/scene relationship. However, these patterns may be neglected by the implicit representation during the learning process. Thus, in the near future, we would like to investigate a more controllable way to update the visual representation.  %\brev{we plan to study more data modalities for fusion, providing richer information for the navigation task in challenging cases.} 
Additionally, complex occlusion relations in the real scenarios often lead to inaccurate height mappings in the occluded areas. In the future, we would like to further utilize the uncertainty map of the environment, together with the multi-view information to improve both the accuracy and the efficiency of the mapping process.
\rev{Moreover, since the geometric modeling (triangulation of sparse keypoints) is commonly involved in the optimization pipeline of SLAM, effectively collaborating the 3D keypoints detection and the height mapping would be quite interesting to explore.}

\section*{Acknowledgment}
We would like to thank the anonymous reviewers for their constructive comments. This work was supported in parts by NSFC Key Project (U2001206), Guangdong Outstanding Talent Program (2019JC05X328), Guangdong Science and Technology Program (2020A0505100064, 2018A030310441, 2015A030312015), DEGP Key Project (2018KZDXM058), Shenzhen Science and Technology Program (RCJC20200714114435012), and Guangdong Laboratory of Artificial Intelligence and Digital Economy (Shenzhen University).

{\small
	\bibliographystyle{cvm}
	\bibliography{VGFNet}

\begin{thebibliography}{10}\itemsep=-1pt

\bibitem{armeni20163d}
I.~Armeni, O.~Sener, A.~R. Zamir, H.~Jiang, I.~Brilakis, M.~Fischer, and
  S.~Savarese.
\newblock {3D} semantic parsing of large-scale indoor spaces.
\newblock In {\em Proc. IEEE Conf. on Computer Vision \& Pattern Recognition},
  pages 1534--1543, 2016.

\bibitem{bansal2019combining}
S.~Bansal, V.~Tolani, S.~Gupta, J.~Malik, and C.~Tomlin.
\newblock Combining optimal control and learning for visual navigation in novel
  environments.
\newblock In {\em Proc. Conf. on Robot Learning}, volume 100, pages 420--429,
  2020.

\bibitem{bian2019unsupervised}
J.~Bian, Z.~Li, N.~Wang, H.~Zhan, C.~Shen, M.-M. Cheng, and I.~Reid.
\newblock Unsupervised scale-consistent depth and ego-motion learning from
  monocular video.
\newblock In {\em Proc. of Advances in Neural Information Processing Systems},
  pages 35--45, 2019.

\bibitem{chaurasia2013depth}
G.~Chaurasia, S.~Duchene, O.~Sorkine-Hornung, and G.~Drettakis.
\newblock Depth synthesis and local warps for plausible image-based navigation.
\newblock {\em ACM Trans. on Graphics}, 32(3):30:1--30:12, 2013.

\bibitem{chen2019learning}
D.~Chen, B.~Zhou, V.~Koltun, and P.~Kr{\"a}henb{\"u}hl.
\newblock Learning by cheating.
\newblock In {\em Proc. Conf. on Robot Learning}, volume 100, pages 66--75,
  2019.

\bibitem{chen2019behavioral}
K.~Chen, J.~P. de~Vicente, G.~Sepulveda, F.~Xia, A.~Soto, M.~V{\'a}zquez, and
  S.~Savarese.
\newblock A behavioral approach to visual navigation with graph localization
  networks.
\newblock In {\em Proc. of Robotics: Science and Systems}, pages 1--10, 2019.

\bibitem{engel2017direct}
J.~Engel, V.~Koltun, and D.~Cremers.
\newblock Direct sparse odometry.
\newblock {\em IEEE Trans. Pattern Analysis \& Machine Intelligence},
  40(3):611--625, 2017.

\bibitem{engel2014lsd}
J.~Engel, T.~Sch{\"o}ps, and D.~Cremers.
\newblock Lsd-slam: Large-scale direct monocular slam.
\newblock In {\em Proc. Euro. Conf. on Computer Vision}, pages 834--849, 2014.

\bibitem{godard2017unsupervised}
C.~Godard, O.~Mac~Aodha, and G.~J. Brostow.
\newblock Unsupervised monocular depth estimation with left-right consistency.
\newblock In {\em Proc. IEEE Conf. on Computer Vision \& Pattern Recognition},
  pages 270--279, 2017.

\bibitem{gupta2017cognitive}
S.~Gupta, J.~Davidson, S.~Levine, R.~Sukthankar, and J.~Malik.
\newblock Cognitive mapping and planning for visual navigation.
\newblock In {\em Proc. IEEE Conf. on Computer Vision \& Pattern Recognition},
  pages 2616--2625, 2017.

\bibitem{henriques2018mapnet}
J.~F. Henriques and A.~Vedaldi.
\newblock Mapnet: An allocentric spatial memory for mapping environments.
\newblock In {\em Proc. IEEE Conf. on Computer Vision \& Pattern Recognition},
  pages 8476--8484, 2018.

\bibitem{huang2020clustervo}
J.~Huang, S.~Yang, T.-J. Mu, and S.-M. Hu.
\newblock Clustervo: Clustering moving instances and estimating visual odometry
  for self and surroundings.
\newblock In {\em Proc. IEEE Conf. on Computer Vision \& Pattern Recognition},
  pages 2165--2174, 2020.

\bibitem{ClusterSLAM}
J.~Huang, S.~Yang, Z.~Zhao, Y.-K. Lai, and S.-M. Hu.
\newblock Clusterslam: A slam backend for simultaneous rigid body clustering
  and motion estimation.
\newblock In {\em Proc. Int. Conf. on Computer Vision}, pages 5874--5883, 2019.

\bibitem{kim2015deep}
D.~K. Kim and T.~Chen.
\newblock Deep neural network for real-time autonomous indoor navigation.
\newblock {\em arXiv preprint:1511.04668}, 2015.

\bibitem{koenig2002d}
S.~Koenig and M.~Likhachev.
\newblock D* lite.
\newblock In {\em Proc. of Association for the Advancement of Artificial
  Intelligence}, pages 476--483, 2002.

\bibitem{kuznietsov2017semi}
Y.~Kuznietsov, J.~Stuckler, and B.~Leibe.
\newblock Semi-supervised deep learning for monocular depth map prediction.
\newblock In {\em Proc. IEEE Conf. on Computer Vision \& Pattern Recognition},
  pages 6647--6655, 2017.

\bibitem{loquercio2018dronet}
A.~Loquercio, A.~I. Maqueda, C.~R. Del-Blanco, and D.~Scaramuzza.
\newblock Dronet: Learning to fly by driving.
\newblock {\em IEEE Robotics and Automation Letters}, 3(2):1088--1095, 2018.

\bibitem{ma2018sparse}
F.~Ma and S.~Karaman.
\newblock Sparse-to-dense: Depth prediction from sparse depth samples and a
  single image.
\newblock In {\em Proc. IEEE Int. Conf. on Robotics \& Automation}, pages 1--8,
  2018.

\bibitem{mirowski2018learning}
P.~Mirowski, M.~Grimes, M.~Malinowski, K.~M. Hermann, K.~Anderson,
  D.~Teplyashin, K.~Simonyan, A.~Zisserman, R.~Hadsell, et~al.
\newblock Learning to navigate in cities without a map.
\newblock In {\em Proc. of Advances in Neural Information Processing Systems},
  pages 2419--2430, 2018.

\bibitem{mur2017orb}
R.~Mur-Artal and J.~D. Tard{\'o}s.
\newblock Orb-slam2: An open-source slam system for monocular, stereo, and
  rgb-d cameras.
\newblock {\em IEEE Trans. on Robotics}, 33(5):1255--1262, 2017.

\bibitem{padhy2018deep}
R.~P. Padhy, S.~Verma, S.~Ahmad, S.~K. Choudhury, and P.~K. Sa.
\newblock Deep neural network for autonomous {UAV} navigation in indoor
  corridor environments.
\newblock {\em Procedia Computer Science}, 133:643--650, 2018.

\bibitem{savinov2018semi}
N.~Savinov, A.~Dosovitskiy, and V.~Koltun.
\newblock Semi-parametric topological memory for navigation.
\newblock In {\em Proc. Int. Conf. on Learning Representations}, pages 1--16,
  2018.

\bibitem{tai2017virtual}
L.~Tai, G.~Paolo, and M.~Liu.
\newblock Virtual-to-real deep reinforcement learning: Continuous control of
  mobile robots for mapless navigation.
\newblock In {\em Proc. IEEE Int. Conf. on Intelligent Robots \& Systems},
  pages 31--36, 2017.

\bibitem{tatarchenko2019single}
M.~Tatarchenko, S.~R. Richter, R.~Ranftl, Z.~Li, V.~Koltun, and T.~Brox.
\newblock What do single-view {3D} reconstruction networks learn?
\newblock In {\em Proc. IEEE Conf. on Computer Vision \& Pattern Recognition},
  pages 3405--3414, 2019.

\bibitem{tateno2017cnn}
K.~Tateno, F.~Tombari, I.~Laina, and N.~Navab.
\newblock Cnn-slam: Real-time dense monocular slam with learned depth
  prediction.
\newblock In {\em Proc. IEEE Conf. on Computer Vision \& Pattern Recognition},
  pages 6243--6252, 2017.

\bibitem{wang2019autonomous}
C.~Wang, J.~Wang, Y.~Shen, and X.~Zhang.
\newblock Autonomous navigation of uavs in large-scale complex environments: A
  deep reinforcement learning approach.
\newblock {\em IEEE Trans. on Vehicular Technology}, 68(3):2124--2136, 2019.

\bibitem{wang2017stereo}
R.~Wang, M.~Schworer, and D.~Cremers.
\newblock Stereo dso: Large-scale direct sparse visual odometry with stereo
  cameras.
\newblock In {\em Proc. Int. Conf. on Computer Vision}, pages 3903--3911, 2017.

\bibitem{piecewiseWang}
W.~Wang, W.~Gao, and Z.~hu.
\newblock Effectively modeling piecewise planar urban scenes based on structure
  priors and cnn.
\newblock {\em Science China Information Sciences}, 62:1869--1919, 2019.

\bibitem{zhu2017target}
Y.~Zhu, R.~Mottaghi, E.~Kolve, J.~J. Lim, A.~Gupta, L.~Fei-Fei, and A.~Farhadi.
\newblock Target-driven visual navigation in indoor scenes using deep
  reinforcement learning.
\newblock In {\em Proc. IEEE Int. Conf. on Robotics \& Automation}, pages
  3357--3364, 2017.

\end{thebibliography}
}

\end{document}